\documentclass[autostyle=false,style=english
]{ceurart}

\usepackage{listings}
\usepackage[disable]{todonotes}
\usepackage{cleveref}
\usepackage{csquotes}
\MakeOuterQuote{"}

\begin{document}

\begin{picture}(0,0)
  \put( \dimexpr(\textwidth+7mm), -0.5\textheight){%
      \rotatebox[origin=c]{90}{%
        \LARGE peer reviewed publication at \href{https://text2sparql.aksw.org/}{Text2SPARQL Workshop @ ESWC 2025}
  }}
\end{picture}

\copyrightyear{2025}
\copyrightclause{Copyright for this paper by its authors.
  Use permitted under Creative Commons License Attribution 4.0
  International (CC BY 4.0).}

\conference{First International TEXT2SPARQL Challenge, Co-Located with Text2KG at ESWC25, June 01, 2025, Portorož, Slovenia.}

\title{ARUQULA - An LLM based Text2SPARQL Approach using ReAct and Knowledge Graph Exploration Utilities}

\author[A]{Felix Brei}[%
orcid=0009-0008-5245-6655,
email=brei@infai.org,
]\cormark[1]\fnmark[1]
\author[A]{Lorenz Bühmann}[%
email=buehmann@informatik.uni-leipzig.de,
]\cormark[1]\fnmark[1]
\author[A,B]{Johannes Frey}[%
orcid=0000-0003-3127-0815,
email=frey@informatik.uni-leipzig.de,
]\cormark[1]\fnmark[1]
\author[A]{Daniel Gerber}[%
]\fnmark[1]
\author[A,C]{Lars-Peter Meyer}[%
orcid=0000-0001-5260-5181,
email=LPMeyer@infai.org,
]\cormark[1]\fnmark[1]
\author[A]{Claus Stadler}[%
orcid=0000-0001-9948-6458,
]\fnmark[1]
\author[A]{Kirill Bulert}[%
orcid=0000-0002-1459-3754,
]

\address[A]{Institute for Applied Informatics at Leipzig University, Goerdelerring 9, 04109 Leipzig, Germany}
\address[B]{Leipzig University, Germany}
\address[C]{Chemnitz Technical University, Germany}

\cortext[1]{Corresponding author.}
\fntext[1]{These authors contributed equally. Authors in alphabetical order}

\begin{abstract}
   Interacting with knowledge graphs can be a daunting task for people without a background in computer science since the query language that is used (SPARQL) has a high barrier of entry.
   Large language models (LLMs) can lower that barrier by providing support in the form of Text2SPARQL translation.
   In this paper we introduce a generalized method based on SPINACH, an LLM backed agent that translates natural language questions to SPARQL queries not in a single shot, but as an iterative process of exploration and execution.
   We describe the overall architecture and reasoning behind our design decisions, and also conduct a thorough analysis of the agent behavior to gain insights into future areas for targeted improvements.
   This work was motivated by the Text2SPARQL challenge, a challenge that was held to facilitate improvements in the Text2SPARQL domain.
\end{abstract}


\maketitle

\section{Introduction}
Knowledge graphs are a modern approach at storing and linking information, that have made their way into several large projects like Wikidata, DBpedia or the Open Research knowledge graph.
Their inherent structure enables the storage of not only the information entities themselves, but also the meaningful relationships between them.
Information from such a graph can be retrieved in various ways, including the use of structured query languages like SPARQL, Cypher, or GQL, as well as alternative methods such as faceted browsing.
While SPARQL is a powerful tool widely used in the semantic web community, it can be challenging for those without training in these technologies.
The last couple of years have shown that large language models (LLMs) can be helpful in translating the intent of a user into a matching SPARQL query, but their precision is still very limited.
To facilitate further research in this area, a contest was created by eccenca GmbH where any individual or group could submit a URL that would return a SPARQL query for a given question (and dataset that the question relates to).
In this paper, we describe the details of our submission to this contest.

\subsection{The TEXT2SPARQL Challenge}
\label{subsec:T2S-Challenge}
The first \textsc{TEXT2SPARQL} Challenge\footnote{\url{https://text2sparql.aksw.org/challenge/}} was designed to evaluate systems capable of translating natural language questions into SPARQL queries across multiple datasets and languages.  
Participants were required to deploy their solutions as publicly accessible RESTful web services.  
Each registered system had to expose a uniform API interface accepting two GET parameters: \texttt{question} (the input in natural language) and \texttt{dataset} (a URL identifying the target knowledge graph).  
The service was expected to return a valid SPARQL query string in response, within a timeout limit of ten minutes.  

The evaluation infrastructure orchestrated 250 queries per endpoint during a 5-day testing phase. 

Two datasets\footnote{\url{https://web.archive.org/web/20250626113401/https://text2sparql.aksw.org/assets/talks/1-sebastian-tramp-introduction.pdf}} were employed to test different aspects of system performance:  
\begin{itemize}
    \item \textbf{DBpedia (DB25)} An English and Spanish subset of DBpedia 2015-10 - a large-scale, multilingual knowledge graph derived from Wikipedia - encompassing 200 selected questions equally split between English and Spanish.  
    The challenge organizers state that they manually curated and validated question-query pairs, with refinements such as \texttt{GROUP BY} and \texttt{ORDER BY} clauses to ensure structural diversity.  
    \item \textbf{Corporate Knowledge Graph (CK25):} A domain-specific KG built from scratch by the challenge organizers.
    It contained 50 English questions reported to be created manually in consultation with corporate stakeholders to simulate realistic enterprise queries. 
\end{itemize}

The challenge thereby tested scalability and multilingual robustness on open-domain data (DBpedia) while assessing domain adaptation and precision on specialized data (Corporate).

\subsection{Contribution}

A recent approach to tackle Text2SPARQL is the use of large language models as agents that can traverse the knowledge graph, retrieve information from it, and create SPARQL queries.
One such agent is SPINACH, released by Stanford~\cite{Liu2024Spinach}.
While SPINACH is an impressive proof of concept implementation targeted at Wikidata, it cannot be easily configured to support other knowledge graphs.

The major contribution of our work consists of the generalization of SPINACH to RDF graphs and adapting and deploying it for the multilingual and multi-KG setting of the TEXT2SPARQL Challenge.  
To this end, we extended SPINACHs codebase and prompting such that it would use our own knowledge graph exploration utilities instead of Wikidata API endpoints. 
These utilities were realized w.r.t. RDF and OWL standard vocabularies, such that they can be adapted to work with other RDF knowledge graphs and languages as well.
In addition, we conducted a qualitative and quantitative analysis of our approach in the TEXT2SPARQL challenge, combining the evaluation logs of the challenge organizers with ARUQULAs action, observation, and reasoning logs.

\section{Related Work}
Although the TEXT2SPARQL Challenge represents the first edition under this specific name, there exists earlier work addressing the problem of translating natural language questions into SPARQL queries.
This task is related to the 
field Knowledge Graph or Knowledge Base Question Answering (KGQA \& KBQA)~\cite{Hoeffner2017SurveychallengesQuestion,Fu2020SurveyComplexQuestion,Perevalov2024Multilingualquestionanswering} and corresponding benchmarks \& leaderboards.
In the scope of this paper, we focus on LLM-powered SPARQL-based KGQA systems \& Text2SPARQL approaches, as well as relevant benchmarks \& leaderboards.

Since the rise of LLMs several colleagues evaluated the capabilities around KGQA and SPARQL.
Lehmann et al.~\cite{Lehmann2023CNL} propose the usage of a controlled natural language as an intermediate step in the Text2SPARQL translation for KGQA.
Meyer et al.~\cite{Meyer2023LLMassistedKnowledge} manually evaluated several KG related capabilities of ChatGPT 3.5 and ChatGPT 4.0, including Text2SPARQL.
The LLMs were generating syntactically correct SPARQL queries with semantic problems on bigger KGs (Mondial KG).
Kovriguina et al.~\cite{Kovriguina2023SPARQLGENOP} applied SPARQLGEN to evaluate a Text2SPARQL approach on bigger KGs (Bestiary, Wikidata and DBpedia) with low F1 score.
Taffa and Usbeck~\cite{taffa2023leveraging} present a KGQA system that finds similar questions in a dataset and uses the corresponding SPARQL queries for a few shot prompt Text2SPARQL translation on ORKG.
This results in a high F1 score, but was tested only for the SciQA benchmark.
The potential of small language models is evaluated by Brei et al.~\cite{Brei2024LeveragingSmallLanguage}.

Text2SPARQL capabilities are evaluated as well in the LLM-KG-Bench~\cite{Meyer2023DevelopingScalableBenchmark,Frey2023Turtle,Meyer2025LLMKGBench3}.
Here we were able to add several SPARQL SELECT query related tasks and assess the capabilities of more than 40 LLMs~\cite{Frey2024AssessingEvolutionLLM,Meyer2024AssessingSparqlCapabilititesLLM,Meyer2025LLMKGBench3,Heim2025ScalingLawsKgeTasks,Meyer2025EvaluatingLLMsForRdf}.
Even for the best state-of-the-art models the F1 scores vary on different datasets.

For Wikidata Liu et al.~\cite{Liu2024Spinach} present SPINACH.
They apply a ReAct~\cite{yao2023react,Sun2023ThinkGraphDeep} approach on the Wikidata KG for a new benchmark dataset.
They reached a quite good performance and the approach looks promising for us.

There exist several efforts for benchmarking KGQA systems.
The KGQA Leaderboard~\cite{DBLP:conf/lrec/PerevalovYKJ0U22}\footnote{KGQA Leaderboard: \url{https://github.com/kgqa/leaderboard}} collects results for common datasets like LC-QuAD~\cite{Trivedi2017LCQuADAC,dubey2017lcQuad2}, QALD~\cite{Usbeck2023QALD10}.
Other notable datasets are SciQA~\cite{Auer2023SciQAScientificQuestion}, Beastiary~\cite{Kovriguina2023SPARQLGENOP} or SPINACH~\cite{Liu2024Spinach}.
New datasets for a KG can get generated as well~\cite{Brei2024Queryfy}.
The GERBIL~\cite{Usbeck2019Benchmarkingquestionanswering} benchmarking platform helps evaluating KGQA systems on datasets.

\section{ARUQULA Approach, Architecture, and Implementation}

As the results of SPINACH~\cite{Liu2024Spinach} and the ReAct approach looked quite promising to us, we decided to generalize the existing code to the RDF graphs given in the Text2SPARQL challenge.
The system is build around a ReAct~\cite{yao2023react,Sun2023ThinkGraphDeep} configuration implemented with LangGraph\footnote{web page: \url{https://www.langchain.com/langgraph}}, a SPARQL endpoint setup realized with RPT\footnote{Repository: \url{https://github.com/Scaseco/RdfProcessingToolkit/tree/master}}  \& Qlever~\cite{Bast2017QLeverQueryEngine}\footnote{Repository: \url{https://github.com/ad-freiburg/qlever}}, a hybrid search realized with the vector database Qdrant\footnote{web page: \url{https://qdrant.tech/qdrant-vector-database/}}, a text search with Lucene, and an API endpoint for the Text2SPARQL challenge.

\subsection{ReAct with KG Exploration Utilities}
\begin{figure}
    \centering
    \includegraphics[width=\linewidth]{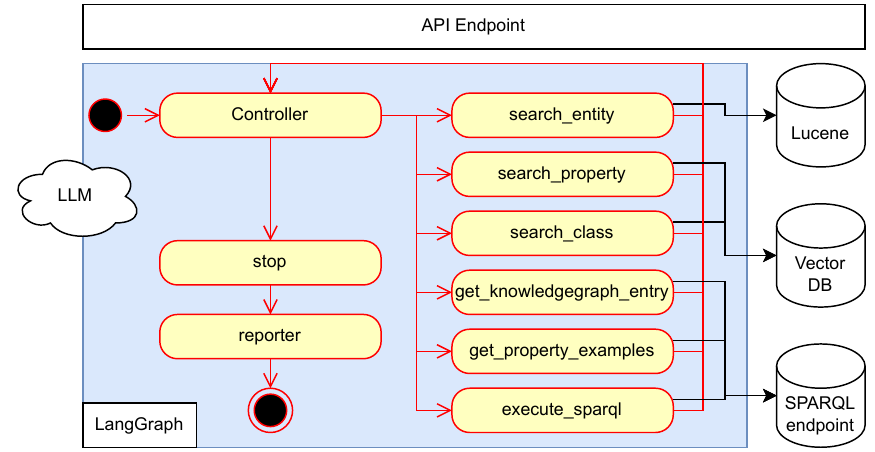}
    \caption{
        Architecture and Action Graph (in the blue area).
        At the controller step the LLM can decide whether to search, inspect, execute or stop.
    }
    \label{fig:architecture_action_graph}
\end{figure}

The ReAct (reason and act) approach~\cite{yao2023react} is built around a graph of actions the LLM can navigate through.
ReAct proposes a template consisting of groups of thoughts, actions and observations which make up the prompt.
The idea here is to not have the LLM try to solve a given task in a single attempt, but instead allow it to split the task into smaller sub-tasks as it sees fit and give it tools to interact with the task as well as a history of all preceding actions and resulting observations.
The available actions and how they interact with the controller (aka. the action graph) are shown in \Cref{fig:architecture_action_graph}.

In the initialization, the language of the query is detected to switch between the English or Spanish DBpedia.
At the "controller" action, the LLM can choose to stop and report the final answer or invoke one out of six knowledge graph exploration utilities:
\begin{description}
    \item[search:] search relevant entities. The \verb|search_entity| action offers a lookup for instance data, \verb|search_property| searches for properties and \verb|search_class| searches for classes.
    The \verb|search_entity| action is implemented as full-text search while the \verb|search_property| and \verb|search_class| actions are implemented with a hybrid vector Search.
    \item[inspect:] get more details on knowledge graph entities. Either with an excerpt on a given entry with \verb|get_knowledgegraph_entry| or some usage examples with \verb|get_property_examples|.
    The action \verb|get_knowledgegraph_entry| is implemented with a SPARQL query searching for outgoing edges.
    The action \verb|get_property_examples| is implemented with a SPARQL query for 5 usage examples for the given property.
    \item[execute:] use \verb|execute_sparql| to test a SPARQL query on the KG.
    The action executes the given SPARQL query on the KG and returns the result.
\end{description}
This actions are described as well in the controller prompt we adopted from SPINACH~\cite{hu2023empirical} as can be seen in \Cref{lst:ActionList}.
The 'controller' step follows after each search/inspect/execute step to select the next action.
This process can be repeated up to 15 iterations which can be changed in the code.

The search/inspect/execute actions all take a single argument.
This could be a string to search for, or an entity to look up or a SPARQL query to execute on the SPARQL endpoint.
In the Python code this actions get translated into function calls with additional parameters like the name of the KG to use or the language detected in the initialization.

\begin{lstlisting}[float,label=lst:ActionList,caption=Action description from the controller prompt, escapeinside={(*@}{@*)}]
- get_knowledgegraph_entry(entity URI): (*@ Retrieves all outgoing edges (linked entities, properties) of a specified knowledge graph entity using its full URI. Example: `http://dbpedia.org/resource/Sufism`. @*)

- search_entity_by_label(string): (*@ Searches the $\{{dataset}\}$ knowledge graph for individual real-world entities like companies, people, locations, or things (e.g. "Apple", "Sufism", "Barack Obama"). @*)

- search_property_by_label(string): (*@ Searches the $\{{dataset}\}$ knowledge graph for *properties* (also called predicates or relationships) like "price", "hasLocation", or "producedBy". Use this when you're trying to find the right property to complete a triple. @*)

- search_class_by_label(string): (*@ Searches for *classes* (types/categories) in the knowledge graph like "Company", "Service", "Book", or "Organization". @*)

- get_property_examples(property URI): (*@ Retrieves a few usage examples of the specified property, given as a full URI. @*)

- execute_sparql(SPARQL): (*@ Executes a SPARQL query on the $\{{dataset}\}$ knowledge graph. Use this when you're confident in your query structure and ready to test a hypothesis. @*)

- stop(): (*@ Marks the most recent SPARQL query as your final answer and ends the process. @*)
\end{lstlisting}

The LLM used needs to have tool support to interact with LangGraph and the tools.
After some internal evaluations of various LLMs including Llama, DeepSeek and various GPT variants we decided to use GPT 4.1 mini as the LLM with the best cost-result-ratio for our case.

\subsection{A Dual-Strategy Approach for Semantic Grounding} 
\label{subsec:dual-strategy}

A critical challenge for any natural language to SPARQL system is semantic grounding: the process of accurately mapping ambiguous or varied natural language phrases from a user's question to the precise, canonical IRIs of classes and properties in the knowledge graph's schema.
This process involves two distinct sub-tasks: resolving conceptual terms (e.g., \emph{"population"}, \emph{"who made this"}) to schema elements (instances of type \texttt{owl:Class}, \texttt{owl:ObjectProperty}, \texttt{owl:DatatypeProperty}), and resolving proper nouns (e.g., \emph{"Berlin"}, \emph{"Google"}) to specific entity instances.
Our agent employs a tailored, dual-strategy approach, recognizing that these two tasks have different requirements for precision and semantic nuance.

\paragraph{Strategy 1: Hybrid Vector Search for Schema Entities}

For grounding conceptual terms against the KG schema, where semantic ambiguity is high, we use the sophisticated hybrid search method, natively supported by the Qdrant vector store\footnote{Qdrant hybrid search query docs: \url{https://qdrant.tech/documentation/concepts/hybrid-queries/}}.
This approach combines dense and lexical search to provide a deep understanding of the user's intent.

\begin{enumerate}
    \item Schema Indexing: We first create a searchable index in Qdrant of all schema entities (instances of type \texttt{owl:Class}, \texttt{owl:ObjectProperty}, \texttt{owl:DatatypeProperty}).
    For each entity, we concatenate its \texttt{rdfs:label} and \texttt{rdfs:comment} into a single text document.
    This document is then encoded into two distinct vector representations:
    \begin{itemize}
        \item Dense Vector: A transformer model (BGE Large English\footnote{Qdrant BGE large-en model: \url{https://huggingface.co/Qdrant/bge-large-en-v1.5-onnx}}) generates a dense embedding that captures the semantic meaning of the entity. 
        This allows for matching based on conceptual similarity.
        \item Sparse Vector: A BM25 based model\footnote{Qdrant BM25 sparse embedding model: \url{https://huggingface.co/Qdrant/bm25}}generates a sparse, high-dimensional vector that excels at keyword-centric matching, ensuring lexical precision for technical or domain-specific terms.
    \end{itemize}
    Both vectors are stored in a Qdrant collection, indexed by the entity's IRI.
    In addition, we also store the domain and range of properties (if available) in metadata fields of the collection.
    \item Agentic Workflow for Schema Grounding: When the agent needs to resolve a term like \emph{"population"}, it generates a dense vector and performs a hybrid query using Reciprocal Rank Fusion (RRF). This robustly identifies the correct schema element (e.g., \texttt{dbo:populationTotal}) by balancing semantic relevance with keyword accuracy.
\end{enumerate}

\paragraph{Strategy 2: Full-Text Search for Named Entity Resolution}

For grounding named entities, the challenge is less about conceptual ambiguity and more about efficiently matching strings against a massive set of instances. 
For this task, a pragmatic and highly performant full-text search is more appropriate.

\begin{enumerate}
    \item Instance Indexing: We use a standard Lucene index, a mature and powerful full-text search library.
    All entity instances from the knowledge graph are indexed. 
    The indexed document for each entity includes its name (\texttt{rdfs:label}) and description (\texttt{rdfs:comment}) (if available).
    \item Agentic Workflow for Named Entity Resolution: 
    When the agent's LLM called in the controller node extracts a proper noun like "Berlin," it does not use the vector store. 
    Instead, it queries the Lucene index. 
    This provides a fast, scalable, and lexically precise method for resolving \emph{"Berlin"} to its canonical IRI, \texttt{dbr:Berlin}.
\end{enumerate}

By employing this dual strategy, our agent effectively uses the right tool for the right job. 
It leverages the semantic depth of hybrid vector search for the nuanced task of schema mapping, while relying on the speed and lexical precision of Lucene for the high-volume task of named entity resolution.
This division is demonstrated when parsing \emph{"What is the population of Berlin?"}: the agent uses hybrid search to ground \emph{"population"} to \texttt{dbo:populationTotal} and Lucene to ground \emph{"Berlin"} to \texttt{dbr:Berlin}, obtaining both components needed to construct the final query.

\subsection{KG Setup \& Querying}
In order to realize the inspection and execution functions, we loaded the CK25 and DB25 KGs into a dedicated SPARQL endpoint each (without making use of named graphs).
In order to make ARUQULA's RDF data management not only conform to the FAIR principles\footnote{FAIR = findable, accessible, interoperable and reusable}~\cite{wilkinson2016fair} but also \emph{reproducible} using a conventional build system, we employed our Maven-based data publishing workflow described in~\cite{Stadler2024fair}. 
The key aspects that make the Maven ecosystem interact well with Semantic Web technology are summarized as follows:
\begin{itemize}
\item Maven artifacts are addressed using \emph{Maven coordinates} which can be represented as URNs of patterns \texttt{<urn:mvn:\{artifact\}>}. A Maven coordinate is resolved using a well-defined conversion that derives a relative URL and prepends it with a repository's base URL. Multiple repositories can be configured for artifact lookups. Changing a repository URL does not necessitate any change in the coordinates of the artifacts.
\item Maven builds are extensible using plugins and there already exist many for common tasks, such as for signing artifacts and validating checksums. We created plugins that build and package RDF databases from a set of RDF data dependencies.
\item Maven has native support for deployment to local folders\footnote{\texttt{mvn deploy -D'altDeploymentRepository=snapshot-repo::default::file:./repo-folder'}}. In combination with a generic file server, this can be leveraged as a lightweight approach to (self-)publishing artifacts that does not require maintenance of a dedicated repository system.
\end{itemize}

For Text2SPARQL, the original dataset downloads were available from the website\footnote{\url{https://text2sparql.aksw.org/challenge/\#corporate-knowledge-small-knowledge-graph}}.
We re-published the individual files as Maven artifacts\footnote{\scriptsize{\url{https://maven.aksw.org/archiva/\#artifact-details-download-content~internal/org.aksw.data.text2sparql.2025/dbpedia/1.0.0}}}, which makes it possible to use them as dependencies in a Maven-aware\footnote{There are several build tools that can interact with Maven repositories, such as Gradle, SBT, or bld.} build process. Initially, we used the \texttt{tdb2-maven-plugin}\footnote{\url{https://github.com/Scaseco/tdb2-maven-plugin}} to load the data into an instance of Apache Jena's TDB2 database. While TDB2's performance is sufficient for small datasets, it becomes a bottleneck for larger ones. For this reason, we created the \texttt{qlever-maven-plugin}\footnote{\url{https://github.com/Scaseco/qlever-maven-plugin}}, which features building QLever\footnote{\url{https://github.com/ad-freiburg/qlever}} databases (via Docker). QLever is presently among the fastest RDF engines and supports the processing of billions of triples on conventional hardware. Our Maven plugin abstraction makes it easy to build a database for either system with a ``push of a button'': Regardless whether one uses TDB2 or QLever, an invocation of~\texttt{mvn package} creates a pre-built database archive, whereas~\texttt{mvn deploy} deploys the archive as well as the \texttt{pom.xml} file to the configured repository. For example, the QLever database is available as yet another Maven artifact\footnote{\scriptsize{\url{https://maven.aksw.org/archiva/\#artifact-details-download-content~internal/qlever.org.aksw.data.text2sparql.2025/dbpedia/1.0.0}}}.
Note, that the deployed \texttt{pom.xml} file serves as a historic snapshot that can be downloaded and used to rebuild the database locally at any point in time (provided that the Maven and Docker ecosystems still exist). By default, our plugins place all triples of an RDF dependency into the graph with the artifact's URN. Consequently, the provenance of data in an RDF store can be tracked using the graph name. However, in practice, it is often desirable to merge multiple datasets into a single graph. For this purpose, our plugins also support the specification of the target graph.
The pre-created databases are used in two ways: The self-contained ARUQULA docker image is built by downloading the QLever database archives directly from our Maven repository. Endpoints are also hosted under our public Apache Jena Fuseki setup\footnote{\url{https://copper.coypu.org/\#/dataset/text2sparql-2025-dbpedia-qlever/info}}. The integration of QLever into Fuseki is part of our JenaX project\footnote{\url{https://github.com/Scaseco/jenax}}.
Our Maven plugins have been published to Maven Central and are thus globally available for use in builds. Other relevant approaches that combine Maven and Semantic Web are \emph{OntoMaven}~\cite{paschke2018ontomaven}, which provides plugins for ontology development and management, and the \emph{DataBus} project~\cite{DBLP:conf/mtsr/FreyGHH21}, which features a large data catalogue that reuses Maven concepts.

\subsection{Challenge API}

As stated in subsection \ref{subsec:T2S-Challenge}, all challengers had to provide a uniform API to expose their service to the judges of the challenge. This API was specified via an OpenAPI conforming JSON document found at \url{https://text2sparql.aksw.org/openapi.json} and defines a single route \verb|/text2sparql| which accepts two parameters via \verb|HTTP-GET|: The name of the dataset that should be queried (in this case limited to \verb|https://text2sparql.aksw.org/2025/DBpedia/| and \verb|https://text2sparql.aksw.org/2025/corporate/|) along with one URL-encoded question. Assuming a base URL of \verb|http://example.com|, a valid request might look like this:

\begin{verbatim}
    http://example.com/text2sparql \
        ?dataset=https%3A%2F%2Ftext2sparql.aksw.org%2F2025%2FDBpedia%2F \
        &question=Who%20designed%20the%20Python%20programming%20language%3F
\end{verbatim}

This API was implemented in Python with the Flask framework\footnote{web page: \url{https://flask.palletsprojects.com/en/stable/}}.

\section{Evaluation}

For each request that was sent to our agent, we logged every prompt, response, action taken, results and total runtime.
Since the challenge consisted of sending 250 such requests, we ended up with a wealth of information.

The first statistic we calculated from data extracted from the agent logs was the average runtime needed to generate a SPARQL query along with the average number of steps that the agent needed to reach this goal.
The results can be seen in table \ref{tab:times-and-steps}.
Given that the timeout defined by the challenge holders was ten minutes, we can see that our agent clearly stayed well below this limit the whole time.

From the data we extracted, we can run some statistical analysis. 
We limit ourselves here to questions from \verb|DBpedia-EN| and \verb|Corporate|.

Running a t-test on the number of agent steps between these two datasets reveals indeed a significant difference.
The calculation results in a $t$-value of $-2.734$ and a $p=0.00744$ which supports the hypothesis that the agent performs notably different between these two datasets.

However, repeating this calculation for the runtimes gives us a $t=-1.770$ and $p=0.079$ which hints in the direction that \verb|DBpedia-en| was quicker to be processed than \verb|corporate|, but the evidence is not strong enough to reach that conclusion.

Given that bandwidth and transfer speed between the agent and the SPARQL endpoint is a contributing factor, as well as the processing power of that endpoint to process a SPARQL query, we must assume that there is a lot of noise in the duration data.
Going with the findings for the number of agent steps though, it seems promising to explore this direction further.

The second avenue of analysis we want to explore is the behavior of the agent itself.
Data scientists follow a certain methodology when searching for new findings in large amounts of data, which consists of drawing samples from the data, analysing their properties, trying to automate this process and finally rolling out this process to a larger portion or even all the data at hand.
The functions that were provided to the LLM were created with that process in mind, giving the LLM the opportunity to approach the process of answer extraction the same way a human would.
But so far, there is no data that conclusively shows that an LLM would indeed follow this process.

\begin{figure}
    \label{fig:action-frequencies}
    \centering
    \includegraphics[width=0.9\linewidth]{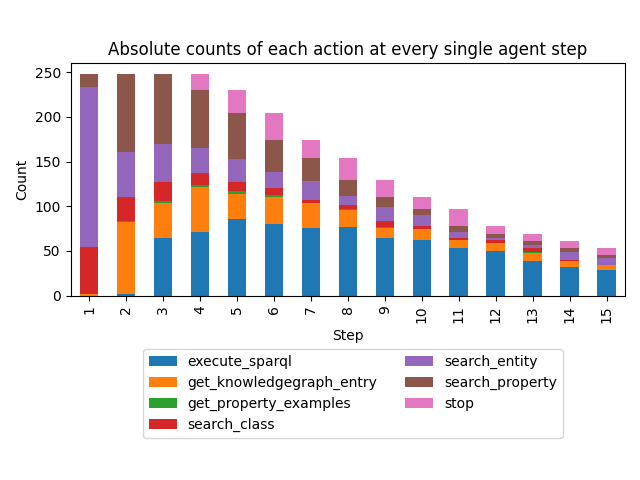}
    \caption{At each individual step of the agent we noted which action was chosen how often from the about 250 queries of the benchmark.
    The step index is shown on the x axis and the height of the bar corresponds to the number of times the agent has reached that step.
    The colors indicate the action selected at a step.}
\end{figure}

Observing the LLM agent trying to answer roughly 250 questions and noting at each step which action was taken how often, we arrive at figure \ref{fig:action-frequencies}.
As first step, the agent mostly considered searching around entities from the graph (\verb|search_entity| and \verb|search_class|).
Step two is mainly concerned with exploring the surroundings of entities by utilizing \verb|search_property| and \verb|get_knowledgegraph_entry|.
In step three and onward, we observed an increasing rate of \verb|execute_sparql|, showing that the agent is ready to send SPARQL queries to the endpoint and expect actionable results.

Figure \ref{fig:action-frequencies} shows the cumulative amount of actions taken up until a certain step index.
Again, we can see that during the first steps there is a steep increase in the total amount of subject related exploration which starts to stagnate a bit after step seven, while the total number of executed SPARQL queries grows at an almost steady pace once the agent is at step three.
The most notable takeaway here is that the \verb|stop| action was called about 200 times by the agent on purpose, meaning that for roughly 50 questions, the agent failed to generate a conclusive SPARQL query and had to scrape one from the conversation trace as a last hail mary.

\begin{figure}
    \label{fig:after-execute}
    \centering
    \includegraphics[width=0.85\linewidth]{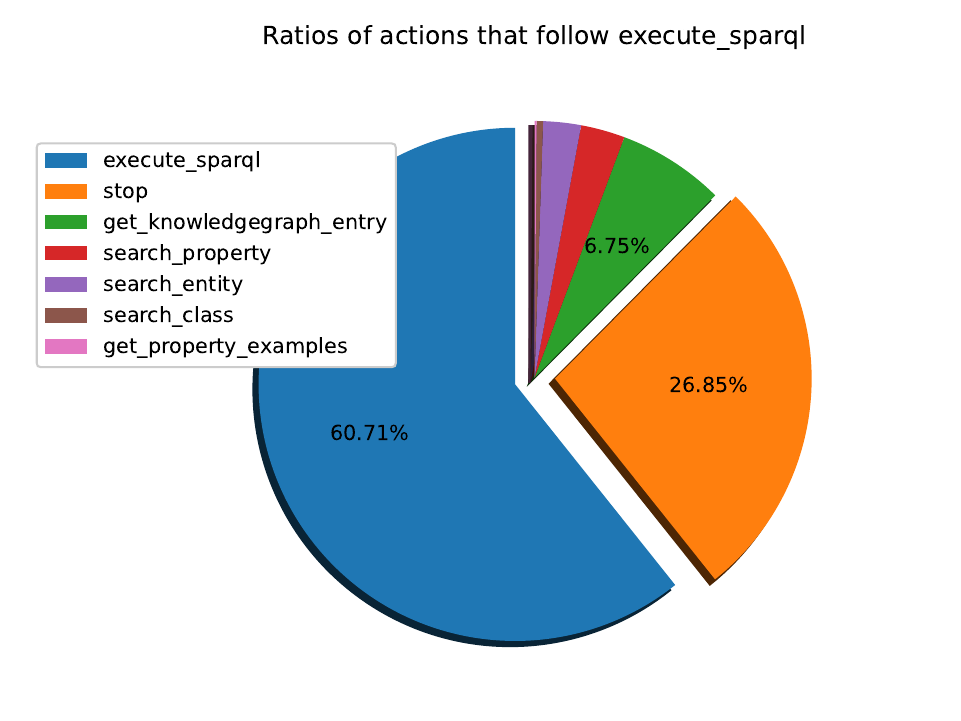}
    \caption{A large portion of actions that the agent takes starting at step three is execute\_sparql. This pie chart shows the ratios of actions that come directly after such an action, i.e. in three out of five times it is followed by yet another execute\_sparql action and in one out of four times the agent decides to stop afterwards, whereas in one out of seven cases the agent resorts to retrieving information from the knowledge graph directly via search or get functions to refine future queries}
\end{figure}

Lastly, we want to investigate further which action precedes which.
To this end, we start by looking at each \verb|stop| action and count the actions that came before.
We find that all 203 \verb|stop| actions were preceded by \verb|execute_sparql| which shows that the agent always verifies its final SPARQL query before issuing a \verb|stop|.
Looking in the other direction, the pie chart in \cref{fig:after-execute} shows the ratio of each action following an \verb|execute_sparql| step.
In one out of four cases, the agent deems the query results satisfying enough to continue with a \verb|stop| action.
However, $60\%$ of all SPARQL executions are followed by another execution immediately.
The reasons for this encompass a swath of different things, mainly empty result sets, syntactic errors and even timeouts from the SPARQL endpoint.
A deeper analysis of why two or more \verb|execute_sparql| actions are chosen consecutively is one goal of our future research.
And lastly, we find that in one out of seven cases the agent refrains from executing SPARQL again and instead returns to searching and inspecting entities and properties of the knowledge graph directly.
The ratios between the agent doing this on its own accord versus the controller forcing the agent to backtrack because it executed the same step twice, is yet another interesting analysis that shall be done in the future.

\begin{table}
    \label{tab:times-and-steps}
    \caption{Average time to generate a SPARQL query from a given question and average number of steps the agent took during generation. The data is grouped by dataset and language. The standard deviation is also shown to illustrate the large spread across the different executions.}
\begin{tabular}{lrrrr}
\toprule
 & \multicolumn{2}{r}{Time to answer (seconds)} & \multicolumn{2}{r}{Number of Steps by agent} \\
 & mean & std & mean & std \\
Name of dataset &  &  &  &  \\
\midrule
corporate & 59.620000 & 25.210858 & 9.920000 & 3.515795 \\
DBpedia-en & 51.440000 & 29.395018 & 8.260000 & 3.486250 \\
DBpedia-es & 66.326531 & 34.449956 & 10.520408 & 3.985100 \\
\bottomrule
\end{tabular}
\end{table}

\begin{figure}
    \centering
    \includegraphics[width=0.85\linewidth]{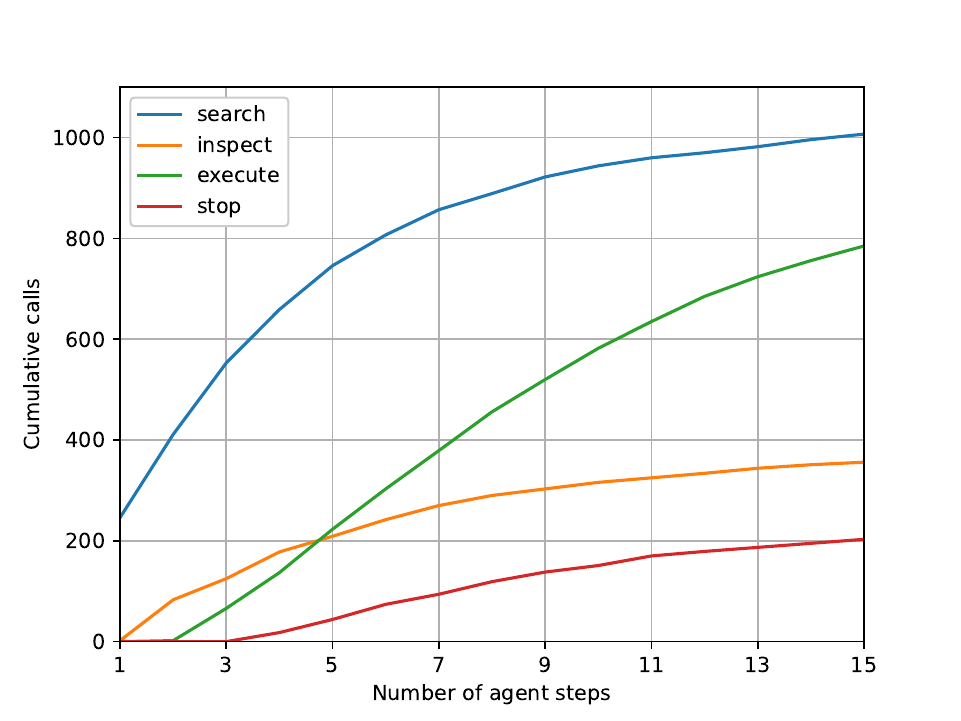}
    \caption{For this plot we grouped the actions into categories to increase readability. It shows the total number of times an action from each category was taken up until a certain step index (i.e. cumulative). For example, we can see that at step seven a total of almost 400 SPARQL queries was executed across all 250 questions. Most notably, only about 200 times total was the stop action called, meaning that circa 50 times the resulting SPARQL query was simply scraped from the conversation trace up until this point without confidence in its correctness.}
    \label{fig:cumulative-actions}
\end{figure}

\section{Discussion}
When comparing our system's query results to the expected results for the DB25 dataset (using the evaluation artifact produced by the challenge organizers), we discovered several pitfalls that need to be considered when interpreting the challenge results and the reported systems` performances\footnote{\url{https://web.archive.org/web/20250627131218/https://text2sparql.aksw.org/assets/talks/8-edgard-marx-result-presentation.pdf}}.

In case of an automatic evaluation of Text2SPARQL performance via SPARQL result set analysis, the query and its result can be viable but do not match with the results of the gold query.
As a result, a low score will be assigned, albeit a good and meaningful query (w.r.t. the question) has been proposed.
On the other hand, a query can achieve a good score w.r.t. its results but in fact it has been constructed in a way that emulates the answer (e.g. using \texttt{VALUES} clauses and \texttt{BIND} statements in combination with internal knowledge of the LLM) without making proper use of the KG.
We could observe instances of misjudged responses of our system for both categories.

When it comes to the first category, the specification of the setup and expected SPARQL query outcome is of significant importance. 
Natural language questions typically lack precise instructions w.r.t. the nature of the SPARQL result projection.
E.g. in case of the question \textit{"What are the 10 most populated countries?"} it is not clear whether the results should return either the entity labels or entity IRIs, or both of them and whether the entities should be reported with the population number, or a further column that specifies the rank. 
While we consider rank and population numbers as not required, given the question, we would also not see providing it as invalid - given the setup instructions of the challenge.
Nevertheless, queries returning labels or population numbers will have a significantly lower score in the challenge.
In terms of the nature of DB25, that consists of DBpedia-EN and DBpedia-ES, it is unclear what kind of entities should be queried and returned. 
Our system was configured in a language-aware way, such that Spanish questions would issue a search for DBpedia-ES entities, but we also saw that the LLM gave preference to Spanish entities during querying by applying Spanish language filters on the entity labels accordingly (thus effectively selecting/querying for DBpedia-ES entities).
However, neither the gold queries nor the challenge setup and specification seem to address this appropriately, leading to scores of 0 for viable queries that return DBpedia-ES entities instead of DBpedia-EN. 

Another problem occurs in case the question can be semantically grounded in several ways for the KG. 
Unfortunately, the DB25 KG provides in several instances multiple options to be queried to answer a particular question.
In case of the population question mentioned above, there are at least 4 different property candidates: \texttt{dbo:population}, \texttt{dbo:populationTotal}, \texttt{dbp:population}, and \texttt{dbp-es:población}. 
Unfortunately, all of them return different results.
While our approach never used \texttt{dbo:population} (which is only used for 7 entities), the Spanish property actually returns factually more accurate results than using the DBpedia Ontology property (which seems to be used in the gold query).
When using the ontology property in combination with the \texttt{dbo:Country} class, the majority of the top-10 results represent associations of countries like the Commonwealth, due to an issue in DBpedia-EN.
Our approach often accounted for such data quality or schema fuzziness issues (that are inherent to the DBpedia extraction and mapping process) and refined the query in that instance such it would e.g. filter out \texttt{dbo:Organisations} in the query.
This poses a significant capability that even outperforms the correctness of the gold query, yet leading to a lower scoring.
We detected several instances where ARUQULA accounted for this with systematic and meaningful constraints on the graph, however, we also found instances where it tried to enforce the "truthfulness" of queries by filtering out result rows with patterns like regex filtering on instance labels or IRIs or faking the outcome with \texttt{BIND} statements and \texttt{VALUE} clauses.
While we consider the enforcement of specific query outcomes an undesirable form of overcompensation in the context of Text2SPARQL, the frequency of this behaviour suggests that both the DB25 dataset and the evaluation setup require refinement to enable precise and reliable automatic evaluation.
Unfortunately, the question reported as running example is only one instance from several problematic question-resultset pairs (some of them even contained false empty result sets).
Nevertheless, we saw a manual analysis of the behaviour and responses of our system w.r.t. DB25 still as an interesting and insightful study.

\section{Future Work}
In this paper, we introduced ARUQULA, a Text2SPARQL agent built upon SPINACH and enhanced with KG exploration utilities to support RDF-based knowledge graphs beyond Wikidata.
Our successful participation in the TEXT2SPARQL challenge demonstrated the potential of the approach, and our evaluation provided insights into agent behaviour, performance characteristics, and limitations.
Looking ahead, there are several avenues to extend and refine our work.
We consider improvements to latency or increased responsiveness of the agent as a requirement to be used within interactive settings, e.g. a chatbot system.
An in-depth comparison of different LLM models in conjunction with a sophisticated selection of test data could help to better understand trade-offs between LLM performance and cost and how the utilities can be enhanced to further improve their helpfulness in order to reduce the number of steps/actions performed. 
Furthermore, it would be interesting to evaluate whether the approach can be transferred to knowledge graphs from other domains and integrate it into evaluation frameworks like LLM-KG-Bench~\cite{Meyer2025LLMKGBench3} or GERBIL~\cite{DBLP:journals/semweb/RoderUN18}.
Thus, it would be beneficial to improve the automation of setup and deployment so that it can be easily configured and deployed for various knowledge graphs.
An evaluation of the ontology grounding by comparing different embedding models, embedding strategies, and search approaches, also in comparison to POTS~\cite{10.1145/3701716.3715194} can be beneficial especially for large knowledge graphs that are not available in the training data of LLMs.

\begin{acknowledgments}
This work was partially supported by grants from the German Federal Ministry of Education and Research (BMBF) to the projects ScaleTrust (16DTM312D) and KupferDigital2 (13XP5230L), as well as from the German Federal Ministry for Economic Affairs and Climate Action (BMWK) to the KISS project (01MK22001A), as well as from the German Federal Ministry of Transport (BMV) to the MobyDex project (19F2266A).
\end{acknowledgments}

\section*{Declaration on Generative AI}
During the preparation of this work, the authors used ChatGPT to: Grammar and spelling check, paraphrase, and reword to improve the writing style.  
After using these tools/services, the authors reviewed and edited the content as needed and take full responsibility for the publication’s content.

\bibliography{ref}

\appendix

\section{Online Resources}
Code Repository: \url{https://github.com/AKSW/ARUQULA/tree/aruqula}

\clearpage
\pagestyle{plain}
\cfoot*{} 

\section*{Metadata for this article}

\begin{description}
    \item[Title:] ARUQULA - An LLM based Text2SPARQL Approach using ReAct and Knowledge Graph Exploration Utilities
    \item[Authors:] Felix Brei* \& Lorenz Bühmann* \& Johannes Frey* \& Daniel Gerber* \& Lars-Peter Meyer* \& Claus Stadler* \& Kirill Bulert \newline
    * These authors contributed equally, authors in alphabetical order
    \item[original publication:] in proceedings of \href{https://text2sparql.aksw.org/}{Text2SPARQL Workshop @ ESWC 2025},
      1.~6.~2025 in Portorož, Slovenia
    \item[submitted for review:] 27.~6.~2025
    \item[peer review status:] accepted by peer review (28.~7.~2025)
    \item[submitted for publication:] 9.~8.~2025
    \item[publication date:] about 2025
    \item[Bibtex entry:] 
\end{description}

\begin{tiny}
  \begin{verbatim}
@InProceedings{Brei2025ARUQULAAnLLM,
  author    = {Brei, Felix and Bühmann, Lorenz  and Frey, Johannes and Gerber, Daniel and Meyer, Lars-Peter and Stadler, Claus and Bulert, Kirill},
  booktitle = {Proceedings of Text2SPARQL Workshop @ ESWC 2025},
  title     = {{ARUQULA} - An {LLM} based {Text2SPARQL} Approach using {ReAct} and Knowledge Graph Exploration Utilities},
  year      = {2025},
}  
  \end{verbatim}
\end{tiny}

\end{document}